\pdfoutput=1

\documentclass[11pt]{article}

\usepackage[]{ACL2023}
\usepackage{multicol}

\usepackage{times}
\usepackage{latexsym}
\usepackage{hyperref}
\usepackage{epsfig}
\usepackage{arydshln}
\usepackage{algorithm}
\usepackage{algorithmic}
\usepackage{booktabs}
\usepackage[T1]{fontenc}

\usepackage[utf8]{inputenc}

\usepackage{microtype}
\usepackage{dsfont}
\usepackage{inconsolata}

\title{Mitigating Catastrophic Forgetting in Task-Incremental Continual Learning with Adaptive Classification Criterion}

\author{
    Yun Luo  \textsuperscript{\rm1 $\#$},
    Xiaotian  Lin  \textsuperscript{\rm2 $\#$},
    Zhen Yang \textsuperscript{\rm3},
    Fandong Meng \textsuperscript{\rm3},
    Jie Zhou  \textsuperscript{\rm3},
    Yue Zhang \textsuperscript{\rm1,4   *}
    \\
    \textsuperscript{1} School of Engineering, Westlake University, Hangzhou, 310024, P.R. China. \\
        \textsuperscript{2} Guangdong University of Foreign Studies. Guangzhou  \\
    \textsuperscript{3} Pattern Recognition Center, WeChat AI, Tencent Inc, Beijing, China.  \\
    \textsuperscript{4} Institute of Advanced Technology, Westlake Institute for Advanced Study, Hangzhou, 310024, P.R. China.  \\
    \texttt{\{luoyun, zhangyue\}@westlake.edu.cn}\\
    \texttt{\{zieenyang, fandongmeng, withtomzhou\}@tentent.com}
}
\begin{document}
\maketitle
\begin{abstract}
Task-incremental continual learning refers to continually training a model in a sequence of tasks while overcoming the problem of catastrophic forgetting (CF). The issue arrives for the reason that the learned representations are forgotten for learning new tasks, and the decision boundary is destructed.  Previous studies mostly consider how to recover the representations of learned tasks. It is seldom considered to adapt the decision boundary for new representations  and in this paper we propose a \textbf{S}upervised \textbf{C}ontrastive learning framework with adaptive classification criterion for \textbf{C}ontinual \textbf{L}earning (SCCL), In our method, a contrastive loss is used to directly learn representations for different tasks and a limited number of data samples are saved as the classification criterion. During inference, the saved data samples are fed into the current model to obtain updated representations, and a k Nearest Neighbour module is used for classification. In this way, the extensible model can solve the learned tasks with adaptive criteria of saved samples.  To mitigate CF, we further use an instance-wise relation distillation regularization term and a memory replay module to maintain the information of previous tasks. Experiments show that SCCL achieves state-of-the-art performance and has a stronger ability to overcome CF compared with the classification baselines. 
\end{abstract}

\section{Introduction}
Continual learning aims to continually train models with new tasks without forgetting previously learned tasks \cite{ke2022continual,De}. It  has become a promising direction for NLP models to incrementally learn new tasks/domains/classes as humans do \cite{ke2022continual}. A typical scenario aims to enable NLP models to solve various tasks in an incremental manner, namely the task-incremental continual learning scenario, which is our study setting in this paper. A salient challenge for continual learning is that continually learned models usually suffer from cartographic forgetting (CF), i.e. the performance on previously learned tasks decreases after training on the new one \cite{lopez2017gradient}.

\begin{figure}
    \centering
    \includegraphics[width=0.8\hsize]{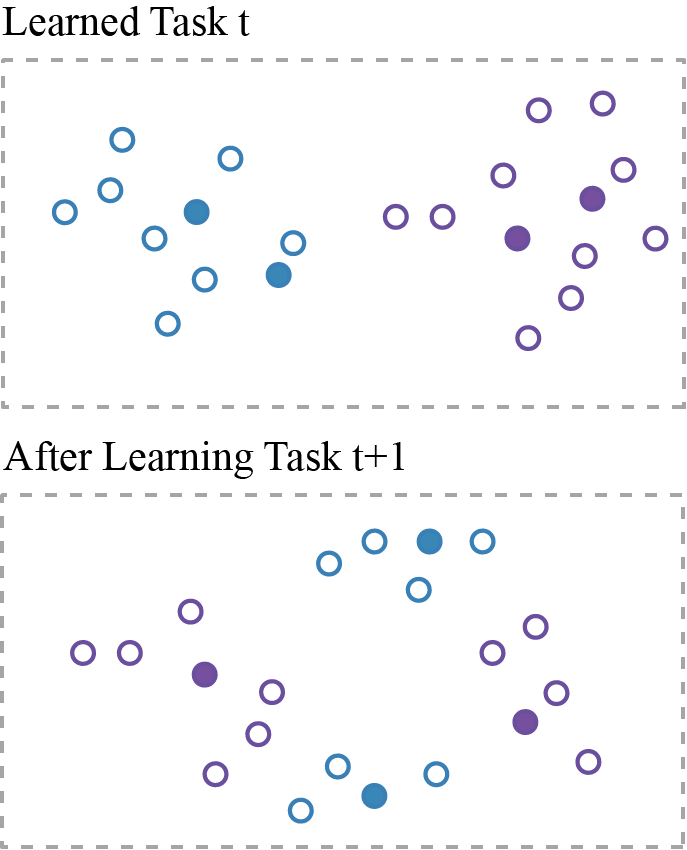}
    \caption{Illustration of representations after contrastive continual learning on a task before and after learning a new task.  }
    \label{illu}
    \vspace{-5mm}
\end{figure}


 Various training strategies have been proposed to mitigate CF \cite{li2017learning,kirkpatrick2017overcoming,lopez2017gradient}. Under the fixed model structure, regularization-based methods design regularization terms to control the shift of representations learned from previous tasks \cite{li2017learning,kirkpatrick2017overcoming,aljundi2018memory}. Rehearsal-based methods  save the data samples from previous tasks into a memory buffer and re-train the model to recover knowledge during training on the current task  \cite{riemer2019learning,lopez2017gradient,de2019episodic}.
 However, most continual learning methods are designed to recover the learned knowledge or mitigate the representation of forgetting. 
 
Seldom considers adapting the classification criterion for the newly learned representations.
For example,  in supervised contrastive learning, the contrastive objective is designed to pull the data representations with the same labels together and push representations with different labels away \cite{Chen2020,gao-etal-2021-simcse,zhang-etal-2021-pairwise,neelakantan2022text,zhaoconsistent}. Representations of the training samples can be saved as a classification criterion, after which an instance-based method such as  a k Nearest Neighbor (kNN) module  can be leveraged for inference \cite{kassnerbert,khandelwal2020nearest}. After learning the new task, we can feed the saved sample into models for new classification criteria and mitigate the problem of CF. For example in Figure 1, although the representations have decayed for learning the new task, the saved samples adapt to serve as the classification criterion in kNN modules.


Inspired by the above motivation, we investigate the use of supervised contrastive learning for task-incremental continual learning (SCCL). After supervised contrastive learning on each task, we use a K-means module to select several samples and save them into a memory buffer while maintaining the  learned representation distribution.  In addition, to mitigate the representation drift when training the model for new tasks, we use an instance-wise relation distillation (IRD) term \cite{fang2020seed,cha2021co2l} and a memory replay module \cite{de2019episodic} to maintain the learned knowledge.
During inference, the saved samples are fed into the trained model to obtain updated representations and a kNN module is used for classification.

Experimental results show that our proposed model can achieve state-of-the-art performance compared with standard cross-entropy-based (CE) baselines. We additionally extend different continual learning strategies \cite{kirkpatrick2017overcoming,aljundi2018memory,li2017learning} to the supervised  contrastive continual learning framework, which gives stronger results than corresponding CE-based methods, showing the advantage of contrastive learning with a kNN classifier in continual learning scenarios. We further analyze  the effectiveness of each module in our paper through ablation studies.  To our knowledge, we are the first to propose a supervised contrastive learning framework for task-incremental continual learning, without any augmented parameters. The code will be released when accepted.

\begin{figure*}
    \centering
    \includegraphics[width=0.9\hsize]{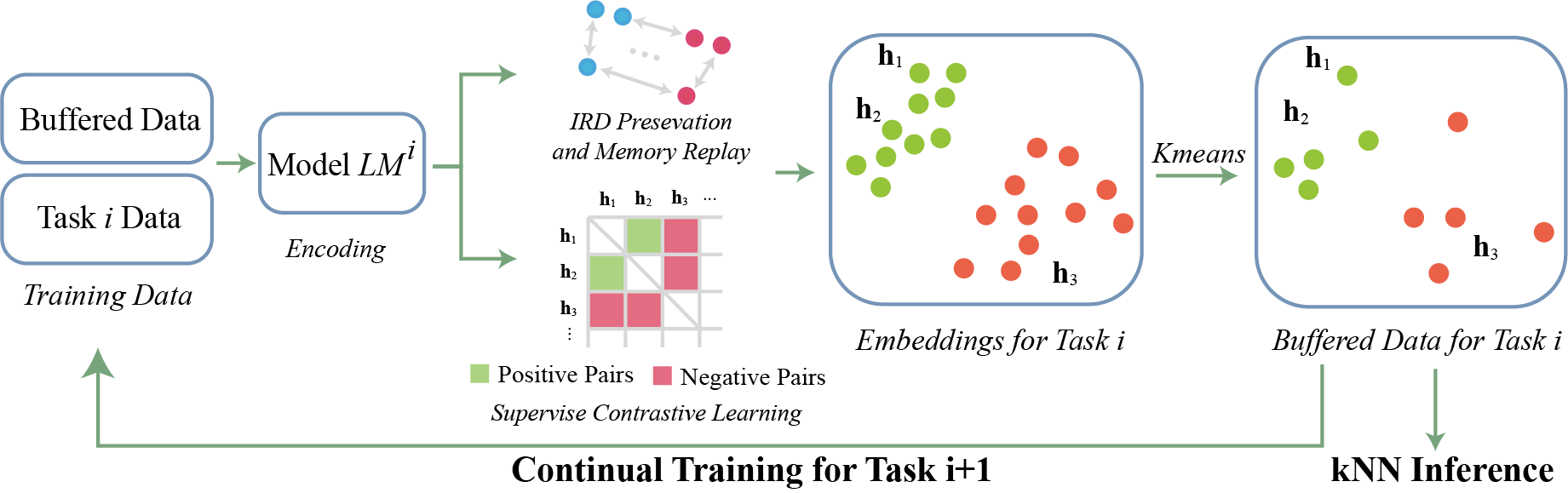}
    \caption{The model framework of SCCL contains four main modules: (1) the supervised contrastive learning for each task; (2) the explicit control of catastrophic forgetting with IRD knowledge distillation and memory replay; (3) the selection of learned representations; (4) a kNN inference module. }
    \label{framework}
        \vspace{-4mm}
\end{figure*}

\section{Related Work}
\textbf{Continual Learning}
Various continual learning methods have been proposed to mitigate the problem of CF. The methods can be broadly divided into architecture-based methods \cite{yoon2018lifelong,serra18a}, regularization-based methods \citet{li2017learning,kirkpatrick2017overcoming}, and rehearsal-based methods \cite{NEURIPS2019_fa7cdfad}.
Under the fixed model structure, regularization-based methods \cite{kirkpatrick2017overcoming,aljundi2018memory,li2017learning} optimize network parameters on the current task while constraining the representation drift. For example, \citet{li2017learning} propose learning without forgetting (LwF) to tackle this problem, which regularizes the model output of current data close to those trained for the previous model. Another category of fixed-structure strategies (rehearsal-based) stores a limited subset of samples from previous tasks to mitigate CF such as ER \cite{NEURIPS2019_fa7cdfad}, RM \cite{Bang_2021_CVPR}, and iCaRL \cite{Rebuffi_2017_CVPR}.

\textbf{Contrastive Learning}
Contrastive learning is initially introduced in self-supervised settings  and proved to  subsume or significantly outperform traditional contrastive losses such as triplet loss \cite{chen2020simple,wu2018unsupervised,gao-etal-2021-simcse,}.  For example, \citet{khosla2020supervised} first propose the idea of self-supervised contrastive learning and prove that the method is more robust to natural corruptions, stable to hyper-parameter settings,  and has strong transfer performance. \citet{luo2022mere} uses supervised contrastive learning combined with a kNN inference module for cross-domain sentiment analysis,  showing a stronger generalization ability compared with standard CE-based methods. 

\citet{cha2021co2l} propose a contrastive continual learning method, Co$^2$L, for class-incremental continual learning. The method uses an asymmetric supervised contrastive loss to enlarge the distance between representations of previous and new tasks.  However, there are significant differences between our model and Co$^2$L. First, the asymmetric contrastive loss of Co$^2$L is unsuitable for task-incremental continual learning,  because a representation can be predicted as different labels according to task objectives. Second, Co$^2$L  uses a decoupled classification layer for inference, i.e. it learns representations first and then learns a linear classifier separately, causing low extensibility and high complexity of the model. In contrast, we use a kNN module as the classification criteria to enhance the extensibility of the model. Third, Co$^2$L only considers the representation drift from the view of regularization. But we also mitigate the problem of representation drift by feeding memory data into the current model to obtain updated classification criteria.

\section{Method}
 The overall SCCL framework is illustrated in Figure \ref{framework}, consisting of four parts. First, we introduce the contrastive learning objective of SCCL in Section 3.1. Second, the selection of learned representations is shown in Section 3.2. Third, an instance-wise distillation module and a  memory replay module are introduced to preserve learned knowledge in Section 3.3 and 3.4, respectively. Fourth, the kNN inference procedure  is shown in  3.5, respectively. The training algorithm is shown in Algorithm 1. 

Formally, a model learns several tasks denoted as $\{T^i\}, i = 1,2,...,n$ ($i$ is the number of tasks).  Each task $T^i$ contains a limited set of labels $C^i$.  During the training of the task $T^i$, only the corresponding data $D^i = \{(x^i_j,y^i_j)\}$ are available, where $x^i_j$ is the input text and $y^i_j \in C^i$ is the corresponding label. In the scenario of task-incremental continual training, the task id can be observed when carrying out inference, and for generality, we consider the label set $C^j\cap  C^k = \emptyset, $ if $i\neq j$.

\subsection{Supervised Contrastive Continual Training (SCCL)}
During the learning on the task $T^i$, we first feed the input $x^i_j$ into a pre-trained language model to obtain hidden states.  The hidden states of a special token $[CLS]$ (the beginning token of the pre-trained language model) are regarded as the representation of the input sequence:
\begin{equation}
    {h}^i_j = Norm(LM^{i}(x^i_j)[CLS]),
\end{equation}
where $Norm(\cdot)$ refers to normalization, $LM^{i}$ is the language model encoder trained for the task $T^{i}$, and $LM^0$ is the initial pre-trained language model.

We denote the data samples in a mini-batch as $A$ (we omit the corner mark $i$ during task $T^i$ for simplicity). For each data sample $j$, we denote $\mathcal{N}(j)\equiv A/\{j\}$, and the positive neighbor set of it as $P(j) = \{u| y_u = y_j  \ and \ u\in N(j)\}$. To push the representations with different labels away, and pull them with the same labels together, we use  supervised contrastive learning objective following \citet{khosla2020supervised}:
\begin{equation}\small
    \mathcal{L}_{cl} = \sum_{j\in A} \frac{-1}{|P(j)|}\sum_{p \in P(j)} log \frac{exp({h}^i_j\cdot{h}^i_p/\kappa)}{\sum_{a\in N(j)}exp({h}^i_j\cdot{h}^i_a/\kappa)}
\end{equation}
where $\kappa$ is the hyper-parameter of temperature.

\subsection{Sample Selection}
After training on each task $T^i$, we select $m$ samples from training data of $D^i$ to keep the representation distribution with respect to the labels (Algorithm 1 (18-22)). In particular, we adopt a K-means module to aggregate the data $D^i(c)$ of each label ($c \in C^i$) to clusters. Then we randomly select samples according to the data density to keep representation distribution, which can be formulated as:
\begin{equation}
    \mathcal{M}^c = Sample(Kmeans(D^i(c)),c,\frac{m}{|C^i|}).
\end{equation}
The selected samples for task $T^i$ are the union of selected data for each label $c$ that $\mathcal{M}^i = \cup_{c \in C^i} \mathcal{M}^c$.  $\mathcal{M}^i$ is saved in the  memory buffer and serves as the classification criteria for task $T^i$  in the continual learning process.


\subsection{Instance-wise Relation Distillation (IRD)}
To preserve the knowledge learned for previous tasks, inspired by \citet{fang2020seed} and \citet{cha2021co2l}, we use an instance-wise relation distillation term to control representation drift (Algorithm 1 (7-9)). During the learning on task $T^i, i>1$, the normalized instance-wise similarity in the mini-batch $A$ is calculated as:
\begin{equation}
    s_{j,p}^i  =\frac{exp({h}^i_j\cdot{h}^i_p/\tau)}{\sum_{a\in N(j)}exp({h}^i_j\cdot{h}^i_a/\tau)},
\end{equation}
where $\mathcal{N}(j)\equiv A/\{j\}$, the representations are encoded by the model $LM^i$ and $\tau$ is the hyper-parameter temperature. Then the IRD regularization term follows: 
\begin{equation}
    \mathcal{L}_{IRD} = \frac{1}{|A|^2}\sum_j\sum_{p} s_{j,p}^{i-1}\ log \ s_{j,p}^{i}.
\end{equation}

The IRD regularization term aims to  estimate the discrepancy of current representations to those learned in the previous model, and mitigate the representation drift through optimization.  In this way, the knowledge of previous  models is preserved and the CF problem can be mitigated.

The overall training objective can be denoted as follows:
\begin{equation}
    \mathcal{L} = \mathcal{L}_{cl}+\mathcal{L}_{IRD}.
\end{equation}

\begin{algorithm}[t]\small
\caption{SCCL Training}
\label{alg}
\begin{algorithmic}[1]
\REQUIRE
A set of training task $\{T^i\}^n$, the corresponding data set $\{D^i\}^n$,  sets of disjoint classes $\{C^i\}^n$. Training steps $S$ and memory replay frequency $f$. Memory buffer size $m$.  Initial pre-trained language model $LM^0$. 
\ENSURE Trained language model encoder $LM^{n}$ and memory buffer $\mathcal{M}$.
\STATE Load pre-trained language model $LM^0$;
\STATE $\mathcal{M} = []$
\FOR{$i = 1,...,n$}
\FOR{$t = 1,...,S$}
\STATE Draw mini-batch $A$ from $D^i$;
\STATE Calculate $\mathcal{L}_{cl}$ of $A$  with $LM^i$ (Eq (1-2));
\IF{$i>1$}
\STATE Calculate $\mathcal{L}_{IRD}$ of $A$  (Eq (5));
\STATE $\mathcal{L} = \mathcal{L}_{cl}+\mathcal{L}_{IRD}$;
  \ELSE 
    \STATE $\mathcal{L} =\mathcal{L}_{cl}$;
\ENDIF
 \STATE Update model parameters with $\mathcal{L}$;
 \IF{$i\ \%\ f == 0$}
 \STATE Update model parameters with memory relay;
\ENDIF
 \ENDFOR
\FOR{$c \in C^i$}
\STATE Obtain k-means clusters of data with label $c$;
\STATE $\mathcal{M}^c = Sample(Kmeans(D^i(c)),c,\frac{m}{|C^i|})$;
\STATE $\mathcal{M}^i=\mathcal{M}^i \cup \mathcal{M}^c$;
  \ENDFOR
\STATE $\mathcal{M}$ = $\mathcal{M} +\mathcal{M}^i $;

\ENDFOR

\end{algorithmic}
\end{algorithm}

\subsection{Memory Replay (MR)}
To make full use of the memory buffer saved during training, we use a memory replay module \cite{de2019episodic} to further recover the knowledge learned in the previous tasks (Algorithm 1 (14-16)). In the training on the task  $T^i, i>1$, we revisit the samples in the memory buffer and train the model with the same loss in Eq (2) after training every $f$ step on the current task.

\subsection{Inference}
 After learning the task $T^i$, we can obtain the model  $LM^i$. During the inference for previous tasks $T^u,\ u<=i$, we feed each test data $x^u_j$ into $LM^i$ and obtain the corresponding representation ${h}^u_j$. Then  we retrieve the $k$ buffered data from  $\mathcal{M}^u$ whose cosine similarity with ${h}^u_j$ is the largest. Note that the representations of buffered data are obtained using the current model, which can adapt to the representation drift  for parameter update.  We denote the $k$ nearest neighbors as ${({h}^u_k,y^u_k)\in \mathcal{K}^u_j}$. The retrieved set is converted to a probability distribution over the labels by applying a softmax with temperature $T$ to the similarity. Using the temperature $T>1$ can flatten the distribution, and prevent over-fitting to the most similar searches \cite{khandelwal2020nearest}. The probability distribution on the labels is expressed as follows:
\begin{equation}
p_{k}(y_j) \propto \sum_{{(h^u_k,y^u_k)\in \mathcal{K}^u_j}} \mathds{1}_{y_j=y^u_k} \cdot exp(\frac{{h}_j^u \cdot {h}_k^u}{T} ),
\end{equation}
and the label with the largest probability is taken as the prediction result.

\section{Experimental Setting}

\subsection{Tasks}
We adopt classification tasks from the benchmark GLUE \cite{wang2018glue} and those from MBPA++ \cite{huang2021continual,de2019episodic}. 
We select dissimilar tasks to form the task sequences, i.e. there are no overlap labels between each task.   The tasks contain 1) CoLA \cite{warstadt2019neural}, requiring the model to  determine whether a sentence is linguistically acceptable; 2) MNLI \cite{williams2017broad} containing 433k sentence pairs annotated with textual entailment information; 3) QNLI\footnote{\href{https://quoradata.quora.com/First-Quora-Dataset-Release-Question-Pairs}{https://quoradata.quora.com/First-Quora-Dataset-Release-Question-Pairs}}, requiring deciding whether the \textit{answer} answers the \textit{question}; 4) QQP, (parsed from SQuAD \cite{rajpurkarsquad}), testing whether a pair of Quora questions are synonymous; 5) Yelp \cite{zhang2015character}, requiring detecting the sentiment of a sentence; 6) AG \cite{zhang2015character}, requiring to classify the topics of the news. 

The sequences can be divided into 2 types with respect to the task lengths: 1) a sequence of 4 classification tasks containing AG, Yelp, QNLI, and MRPC; 2) a sequence of 6 classification tasks containing AG, MRPC, MNLI, CoLA, Yelp, and QNLI.  Without losing generality the orders are randomly selected and the task orders for experiments are shown in Table \ref{seq}.

\begin{table}[] \small
\centering
\begin{tabular}{ll}
\hline
\multicolumn{2}{l}{Orders}         \\ \hline

1      & AG $\rightarrow$ Yelp $\rightarrow$ QNLI$\rightarrow$ MRPC \\
2      & MRPC $\rightarrow$ QNLI $\rightarrow$ Yelp $\rightarrow$AG  \\
3      & QNLI $\rightarrow$Yelp $\rightarrow$MRPC$\rightarrow$AG \\ 
4 & AG$\rightarrow$MRPC $\rightarrow$CoLA$\rightarrow$MNLI $\rightarrow$Yelp$\rightarrow$ QNLI \\
5 & QNLI $\rightarrow$Yelp $\rightarrow$MNLI$\rightarrow$CoLA $\rightarrow$MRPC$\rightarrow$ AG \\
6 & MNLI $\rightarrow$ AG $\rightarrow$QNLI$\rightarrow$ 
 MRPC $\rightarrow$Yelp$\rightarrow$ CoLA \\
\hline
\end{tabular}
\caption{Different  task orders for our experiments.}
\label{seq}
    \vspace{-5mm}
\end{table}

\begin{table*}[]\small
\centering
\begin{tabular}{l|cc|cc|cc|cc|cc|cc}
\hline 
Model  & \multicolumn{2}{c|}{Order 1} & \multicolumn{2}{c|}{Order 2} & \multicolumn{2}{c|}{Order 3} & \multicolumn{2}{c|}{Order 4} & \multicolumn{2}{c}{Order 5}& \multicolumn{2}{c}{Order 6} \\ \hline
       & ACC           & BWT         & ACC          & BWT         & ACC          & BWT         & ACC          & BWT         & ACC          & BWT   & ACC          & BWT       \\
       \hline
Joint & 83.09 & -  &  83.09 & - & 83.09 & -  & 83.06 & - & 83.06 & - & 83.06 & -  \\
\cdashline{1-13}[2pt/4pt]
CE     & 54.74 & -36.96 & 59.43 & -31.11  & 51.73  &  -41.22 & 56.05  & -29.71 & 49.70 & -32.32 & 48.64 &-39.96 \\
CL     &  68.10  & -14.74    & 59.95   & -23.13     &  60.13    &  -22.72    &  62.35   &-24.28   & 58.65    & -27.26  & 63.36  &  -18.7           \\
\cdashline{1-13}[2pt/4pt]
CE-MAS &  59.91    & -23.53   &  61.21
    &  -24.45  &  61.75  & -19.25            &  62.52   & -19.25    & 54.77          &  -17.26  & 58.30 & -32.58
         \\
CL-MAS & 68.99   &   -1.37     &  71.83            &  -3.55    &  73.61   &  -6.94        & 69.95           & -2.89  & 69.27    & -6.27   & 68.93  & -2.44         \\ 
CE-EWC & 66.11      & -22.83     &  71.44            & -14.56     & 69.66    & -16.47            &  62.95            & -22.42    & 59.22     & -25.79  &61.87  &-22.81             \\
CL-EWC &  73.19      &   -0.59     &  75.89            &    -2.04         &  73.52            &   -5.97      &  66.74    & -3.65      & 68.83      &  -6.27  & 68.56  &  -4.00         \\
CE-LwF &  72.09     & -13.26      & 72.13             &  -14.39       & 73.54      &  -12.36       &  68.23     & -13.84     &  63.15    & -22.72  & 67.92  & -17.38            \\
CL-LwF &  76.53     &  -0.33      & 79.15      &  -3.71           &  79.58            &   -3.23          & 68.24            & -5.34   &    72.39     & -9.83    & 71.48  & -5.97        \\

\cdashline{1-13}[2pt/4pt]
CE-ER  &  76.83     & -9.13   & 76.60             & -10.39   & 76.90   &  -12.59           &  75.08    &  -10.05   & {76.51}        & -6.39   & 76.15  & -14.61            \\
IDBR& 75.70  & -3.16  & 73.62  & -7.43  &  75.11  & -3.65 & 65.40 & -10.56 & 69.94 & -5.96  & 66.30  & -10.86\\
Co$^2$L&70.58 & -2.07 & 74.02 & -7.52 & 74.10 & -7.68 & 64.31 & -3.57 & 65.04 & -10.62 & 64.94 &-15.65 \\
\cdashline{1-13}[2pt/4pt]

\textbf{SCCL} & \textbf{79.20} & -2.93 & \textbf{80.05}  & -3.07    &  \textbf{80.24}  &  -3.51  &  \textbf{78.36}   &  0.57   & \textbf{79.00}   & -3.39  & \textbf{78.55}  & -3.75  \\
\ \ {{w/o MR}} & 77.19 & -5.34  & 78.63  &  -4.59   &  80.27  &  -2.91   &  75.65   & -2.91     &  71.22  &  -11.62  &  74.64   & -7.30             \\
\ \ {{w/o IRD}} & 77.57 & -5.33 & 79.73  & -2.48    &  79.48  &  -3.33  &  73.87   &  -6.62   & 76.89   & -4.10  & 74.32  & -4.03   \\
\hline
\end{tabular}
\caption{Continual Learning results on 6 different tasks.  `CE' refers to the standard cross-entropy-based methods, and `CL' refers to extended  contrastive-learning-based methods with continual learning strategies. `-' for not acquirable. All the results are averaged on 5 different random seeds. }
\label{results}
    \vspace{-5mm}
\end{table*}

\subsection{Evaluation Metrics}
We adopt the metrics of average accuracy (ACC) and backward transfer (BWT) to evaluate the performance of the continual learning model \cite{lopez2017gradient}. The model trained after the task $T^i$ is evaluated on the test set of earlier tasks $T^j$ ($j<=i$), and the test accuracy is denoted as $R^i_j$. The metrics are shown as follows:
\begin{equation}
    ACC = \frac{1}{n} \sum_{i=1}^{n} R^n_i
\end{equation}

\begin{equation}
    BWT = \frac{1}{n-1} \sum_{i=1}^{n-1} R^n_i - R^i_i,
\end{equation}
where the former evaluates the overall performance of the final trained model, and the latter calculates the knowledge forgetting during the continual training procedure.


 \subsection{Baselines} We not only compare our model with several CE-based continual learning methods but extend training strategies of them to our contrastive learning framework (i.e. training with contrastive learning and inferring with kNN) to verify the effectiveness of contrastive learning in mitigating CF. We also compare our model with the competitive models IRDB \cite{huang2021continual} and Co$^2$L \cite{cha2021co2l}.  The shared hyper-parameters are kept the same as SCCL in baselines. The model details are as follows:

\begin{itemize}
     \vspace{-2mm}
     \item \noindent \textbf{Fine-tune} (CE, CL) \cite{yogatama2019learning} modifies the parameters of the pre-trained language model to adapt to a new task without any augmented strategies and additional loss.
     \vspace{-2mm}
     \item \noindent \textbf{Experience Replay} (ER) \cite{riemer2019learning} stores a small subset of samples from previous tasks and replays those to prevent models from forgetting past knowledge. 
    \vspace{-2mm}
     \item \noindent \textbf{Elastic Weight Consolidation} (EWC) \cite{kirkpatrick2017overcoming} slows down the updates of the optimal parameters for previous tasks by extending the loss function with a regularization term. 
 \vspace{-2mm} 
     \item \noindent \textbf{Memory Aware Synapses} (MAS) \cite{aljundi2018memory} slows down the update according to the importance weight of each parameter in the network, i.e. the sensitivity of the output function to a parameter change. 
\vspace{-2mm} 
     \item \noindent \textbf{Learning Without Forgetting} (LwF) \cite{li2017learning,9156964}  aims to keep the model output of current data close to those of the previous model. 
  \vspace{-2mm} 
     \item \noindent \textbf{IDBR} \cite{huang2021continual} uses information  disentanglement regularization to encode task-specific information and general information individually, which are jointly considered for classification.
 \vspace{-2mm} 
      \item \noindent \textbf{Co$^2$L} \cite{cha2021co2l} uses an asymmetric supervised contrastive learning method to learn representations and trains  a decoupled  layer for inference.
  \vspace{-2mm}
      \item \noindent \textbf{Multi-task Training} (Joint) \cite{NEURIPS2020_3fe78a8a}  trains on all the tasks simultaneously, i.e. the data of different tasks are mixed up for training. It does not suffer from catastrophic forgetting and represents an upper bound on model performance. 
\end{itemize}

     




\subsection{Implementation Details}
 We adopt the officially released \textit{roberta-base} from HuggingFace \footnote{\href{https://huggingface.co/}{https://huggingface.co/}} as our backbone network.  We train our model on 1 GPU (A100 80G) using the Adam optimizer \cite{kingma2014adam}. For all the models, the batch size is 96,  the learning rate is 3e-5, and the scheduler is set linear. We train our model 10 epochs for each task. Following \citet{huang2021continual}, we select 4,000 samples for each label in training. The hyper-parameters of temperatures $\kappa$ is 0.2, $\tau^*$ is 0.2, and $T$ is 5, and the number of nearest neighbors $k$ is 10. The  memory size for each task is set to 200 (2.5\% of the training data) and the memory replay frequency $f$ is 100.  Through the training of our model, no development set is applied to find the best checkpoints, but stop until the training step is reached.

\section{Results}
\subsection{Overall Results}
The overall results of our experiments are shown in Table \ref{results}. First, our model SCCL achieves ACCs of 79.20\%, 80.05\%, 80.24\%, 78.36\%, 79.00\%, and 78.55\% in Order 1-6, respectively, which are 2.37\%, 0.9\%, 0.66\%, 3.28\%, 2.49\%, and 2.40\% higher than the second-best performance of the continual learning baselines. 
It shows that the performance of the continually learned model is well-maintained in SCCL, but the  problem of CF still exists. SCCL achieves state-of-the-art ACCs compared with the baseline models, indicating the effectiveness of our proposed framework. We also observe that the performance variance is small in the SCCL model for different orders, which implies that our models are not sensitive to the order of task sequences. 

Second, the results of BWT range from -3.75\% to 0.57\% in SCCL for Orders 1-6, which demonstrates knowledge forgetting during the continual learning procedure. The results of SCCL are relatively higher than CE-based models, indicating that SCCL suffers from a milder impact of CF.   Note that  the BWT of SCCL is 0.57\% in  Order 4, which indicates that SCCL can even backward transfer the knowledge from the current tasks to previous tasks. But compared with CL-LwF, CL-MAS, and CL-EWC, the values  of ACCs in SCCL are higher, but BWTs are adverse.  It implies that using the regularization-based strategies, the fine-tuning performance is destructed for explicit control of representations. In this way, BWTs become low since the fine-tuning performance on downstream tasks is relatively weak.

\begin{figure*}
    \centering
    \includegraphics[width=\hsize]{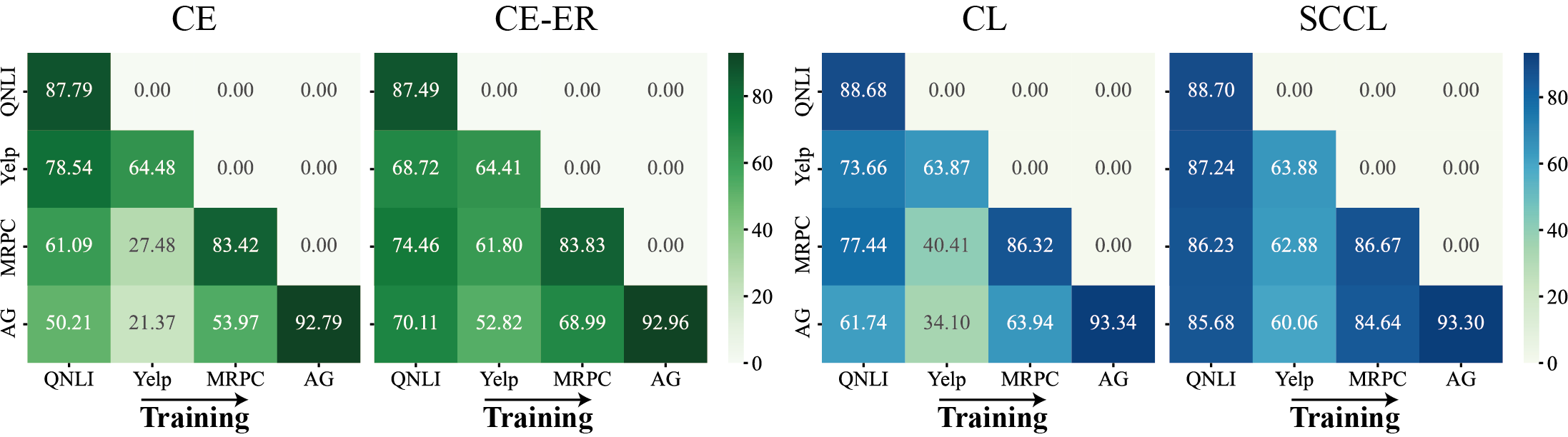}
    \caption{Detailed results during continual learning procedure for different strategies in Order 3.}
    \label{details}
    \vspace{-5mm}
\end{figure*}

\begin{figure}[pt]
    \centering
    \includegraphics[width=0.87\hsize]{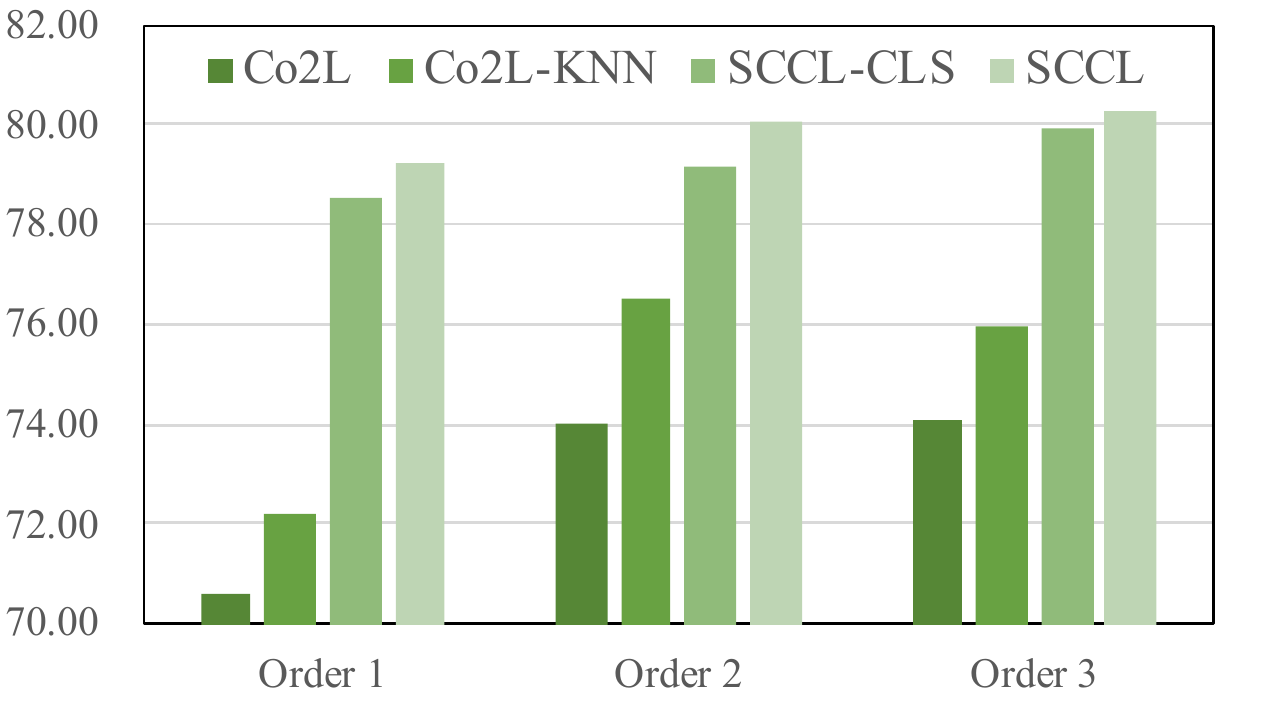}
    \caption{Comparisons of SCCL and Co$^2$L with ablation studies.}
    \label{abl}
    \vspace{-6mm}
\end{figure} 

Third,  the  extended CL-based models achieve stronger performance than corresponding standard CE-based models. For example,  the model CL-LwF achieves 
ACCs of 76.53\%, 79.15\%, 79.58\%, 68.24\%, 72.39\%, and 71.48\%, which are 4.44\%, 7.02\%, 6.04\%, 0.01\%, 9.24\% and 3.56\% higher than those of CE-LwF. The results of CL, CL-MAS, and CL-EWC are in a similar pattern.  The results reflect that contrastive learning with a kNN classifier for continual learning has a stronger ability to overcome CF. But we observe that Co$^2$L achieves relatively low performance compared with our model, which proves that Co$^2$L is not effective for task-incremental learning. It can be explained that Co$^2$L keeps the knowledge of classes and separate the tasks with clear boundary, by using asymmetric supervised contrastive loss, which makes it difficult to distinguish a representation for different task purposes.

Finally, we observe a significant variance in the results of different task orders for regularization-based methods. For example, ACCs of CL-EWC range from 75.89\% to 66.74\%.  But in CE-ER or SCCL the variance is less drastic, such as CE-ER ranging from 76.90\% to 75.08\% and SCCL ranging from 80.24\% to 78.55\%. The phenomenon may result from that knowledge forgetting of previous tasks increases step by step for information los, but no samples help recover such information in regularization-based methods.

\subsection{Ablation Study}
We show the ablation study of memory replay and IRD in the last two rows in Table 2. The ACCs of the models w/o memory replay range from 71.22\%, to 80.27\% for Order 1-6,  which are 2.01\%, 1.42\%, -0.03\%, 2.71\%, 7.78\%, and 3.91\% lower than SCCL, respectively.  It shows the effectiveness of memory replay, without which ACC also becomes less robust to task orders. Then ACCs of the models w/o IRD are 1.63\%, 0.32\%, 0.76\%, 4.49\%, 2.11\%, and 4.23\% lower than SCCL for Order 1-6, respectively. We observe that the models w/o IRD are more robust to task orders, which implies that rehearsal-based methods  are less sensitive to task sequences. Comparing the model w/o IRD with CE-ER, the model performance are also higher than those of CE-ER, which uses almost the same training strategy. The phenomenon demonstrates the effectiveness of contrastive learning in overcoming CF.

We also compare our model with Co$^2$L in ablation studies (Figure \ref{abl}). First, we replace the kNN module of SCCL with a decoupled linear classifier like \cite{cha2021co2l} (SCCL-CLS), where ACCs are slightly smaller than SCCL. It indicates that the kNN module in SCCL can achieve satisfactory performance without additional training on the final representations of contrastive learning. Then we replace the decoupled linear classifier of Co$^2$L with our kNN module (Co$^2$L-kNN), and we observe an increase in performance. It implies that the representations learned by  Co$^2$L are not separated clearly in the feature space, thus a trained linear layer is less effective for classification. But K-means selection of the samples and kNN inference module can estimate the representation distribution more precisely, resulting in better performance. Note that the results of  SCCL are also stronger than Co$^2$L-kNN, which indicates the effectiveness of our model on task-incremental continual learning.

 \begin{figure*}
    \centering
    \includegraphics[width=0.9\hsize]{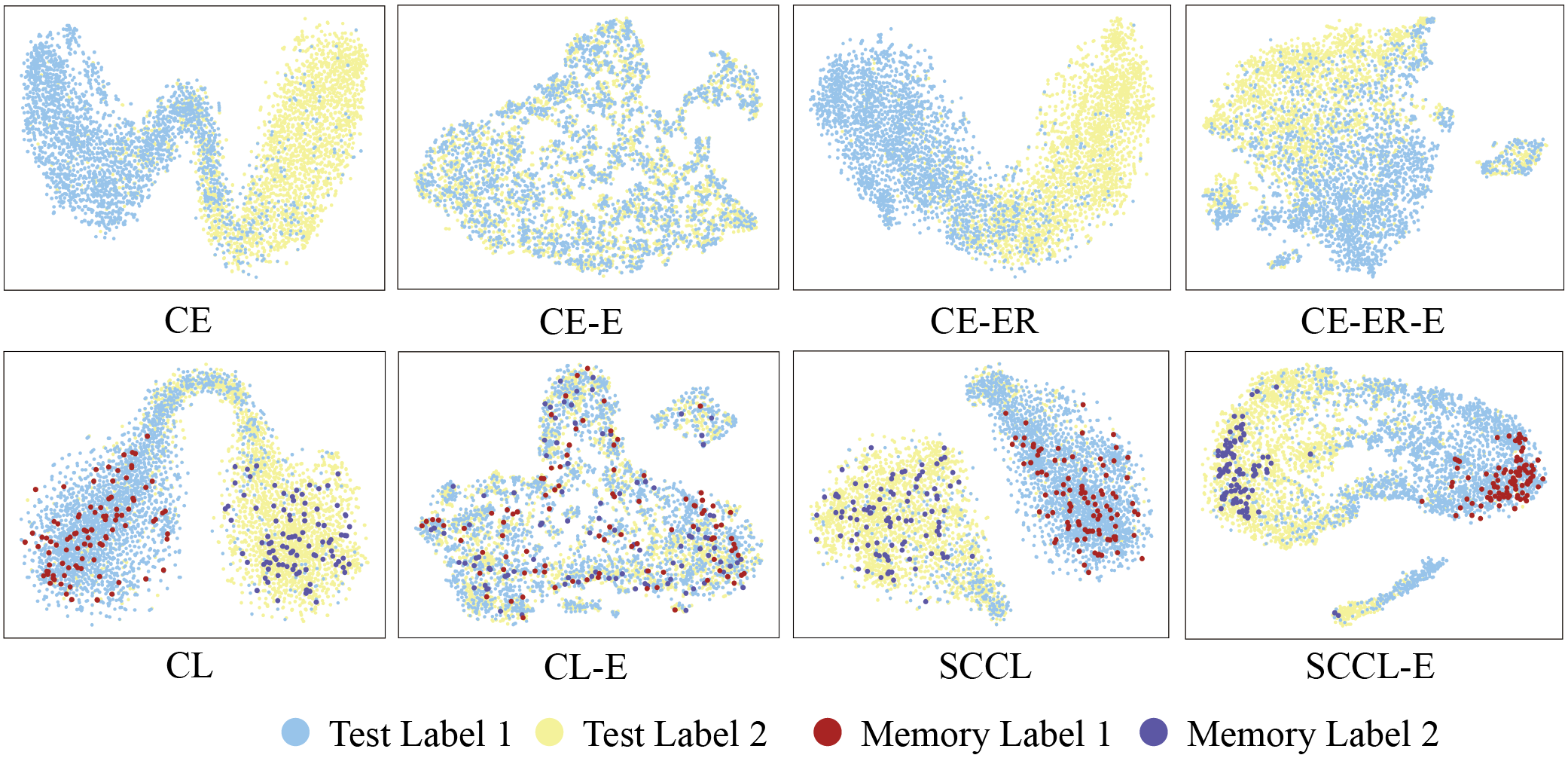}
    \caption{t-SNE visualization of the representations of QNLI samples learned based on the different continual learning methods in Order 5. `E' refers to the representations at the end of continual learning.}
    \label{vis}
        \vspace{-5mm}
\end{figure*}

\subsection{Detailed Results}
As an example, we show the detailed results of Order 3 in several models (Figure \ref{details}). 
First, in the model of CE, we observe that the test accuracies of QNLI decrease from 87.78\% to 50.21\% step by step with the continual training on the task QNLI, Yelp, MRPC, and AG. The accuracies of Yelp and AG are also in a similar pattern, where the final performance is nearly random. It indicates that using standard CE for continual learning suffers from CF significantly. But in the method CL, the final performance of QNLI, Yelp, and MRPC is still stronger than a random prediction, indicating that contrastive learning with the kNN module can maintain learned knowledge in each training step and results in satisfactory performance at the end.

 The model CE-ER can also mitigate CF compared with CE and CL, but the performance still decreases a large margin in the task of QNLI, Yelp, and MRPC. The accuracy of QNLI decreases by 17.38\%, that of Yelp decreases by 11.59\%, and that of MRPC decreases by 14.84\%. 
 As for our model SCCL, we observe that the test performance is 88.70\%, 87.24\%, 
86.24\% and 85.68\% after training on tasks QNLI, Yelp, MRPC, and AG, respectively. It shows that the performance of SCCL decreases as the training precedes, but within a small range (3.02\%). The results on Yelp and MRPC are in a similar pattern. It demonstrates that our model has a strong ability to overcome CF.

\subsection{Visualization}
We use t-SNE to  visualize the representations of QNLI in Order 3 of the training models, CE, CE-ER, CL, and SCCL (Figure \ref{vis}). As we observe in CE the representations of the test data are clearly separated into two clusters after training on the task QNLI. When finishing the continual learning, the representations become nearly uniformly distributed on the feature space and the model only achieves an accuracy of 50.21\%. It demonstrates that catastrophic forgetting is significant due to representation drift. In the model CL, the representations drift severely as well, but the distribution is less uniform compared with CE. Typically, we can clearly at the upper right of the distribution, there are more memory samples with label 1, and the test samples with label 1 also gather in the position, indicating correct classification based on kNN. The test performance achieves 61.74\%, but is still 26.94\% lower than the initial model. The phenomenon shows representations during continual learning drift less significantly and the saved samples (the classification criterion) also drift, which maintains some correct inferences. But CF is still a salient problem in contrastive learning.

But in CE-ER, the boundary of the representations becomes indistinct, and the accuracy of QNLI decreases from 87.79\% to 70.11\% after continual learning. It indicates that the representations are less effective compared with the initially trained, i.e. CF is significant in CE-ER. But the representations in SCCL are  still clearly divided into two parts according to the labels. The representations of the memory samples are among the according clusters, implying the performance on the task QNLI is well-maintained. Correspondingly, the accuracy at the end of learning is 85.68\% based on SCCL, only 3.02\% lower than the initial performance. It shows that in SCCL the representation drift slightly and the classification criterion is well-maintained, resulting in a satisfactory performance.



\section{Conclusion}
In this paper, we proposed a supervised contrastive learning model for task-incremental continual learning (SCCL) to boost the extensibility of continual learning. The model used contrastive learning to learn representations and a kNN module was adopted for inference, together with an instance-wise distillation and a memory replay module to maintain previously learned knowledge.   With extensive experiments, our model achieved state-of-the-art performance compared with standard CE-based methods. Ablation studies and visualizations also proved the effectiveness of our model in solving the problem of CF. 

\clearpage
\section{Limitations}
Our model SCCL is specific for task-incremental continual learning scenarios, but not suitable for class-incremental scenarios. In class-incremental scenarios, the representations of current classes should be designed to be far away from previous ones. For simplicity,  we do not consider data augmentation in our model, so the batch size should be large enough to contain positive pairs for each label. But data augmentation (such as two different dropout representations \cite{gao-etal-2021-simcse}) is a plug-and-play module for our model if there are plenty of labels in each task.

\bibliography{ref}
\bibliographystyle{acl_natbib}

\clearpage
\appendix

\section{Data Statistics}
We show the data statistics in Table \ref{task}.
\begin{table}[h] \small
\centering
\begin{tabular}{ccccc}
\hline
Task &Type &\#Train &\#Test &\#Labels         \\ \hline

 AG &News & 16000 & 7600 & 4 \\
QNLI & Q \& A & 8000 & 5266 & 2\\
Yelp & Sentiment & 20000 & 7600 & 5\\
CoLA &Linguistics  & 6527 & 1042 & 2\\
MNLI & Inference & 12000 & 9815 & 3 \\
MRPC &Paraphrase & 4074 & 1725 & 2\\
\hline
\end{tabular}
\caption{Statistics for different classification tasks.}
\label{task}
    \vspace{-2mm}
\end{table}

\section{kNN Sensitivity}
\begin{figure}[p]
    \includegraphics[width=0.95\hsize]{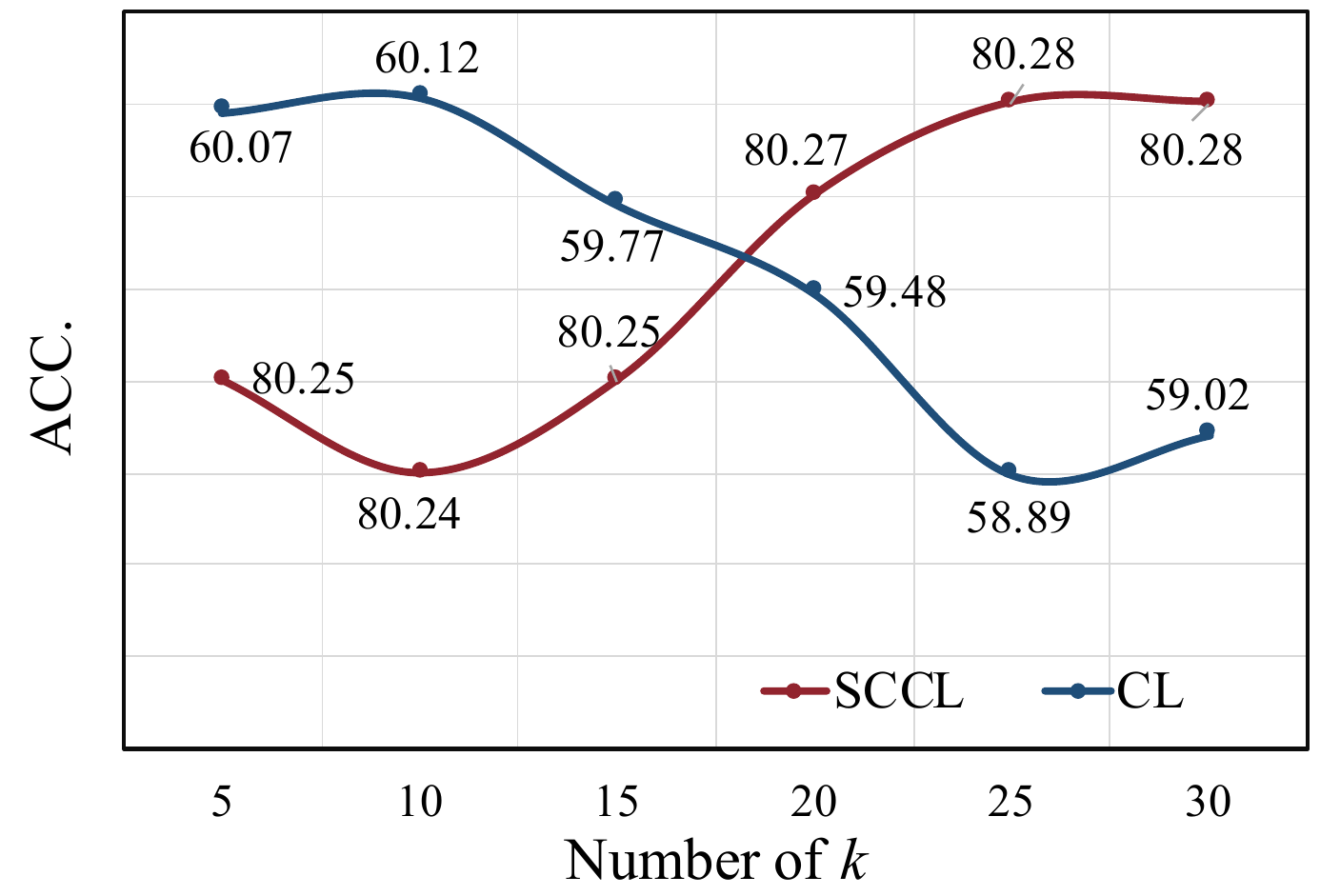}
    \caption{Test results with respect to different numbers of $k$  for Order 3.}
    \label{knn}
    \vspace{-4mm}
\end{figure}

We show the sensitivity of SCCL to the number of $k$ in the kNN module (Figure \ref{knn}). We find that the performance of our model  fluctuates from 80.24\% to  80.28\% (a significantly small range) in our method, indicating  the representations in our model cluster well in the feature space and are robust to the hyper-parameter $k$.   But the performance in CL fluctuates more severely, ranging from 
 59.02\% to 60.12\%.  The best performance is achieved when $k=10$ and decreases with the increase of $k$, which means the representations drift significantly and the clusters become less reliable. The experiment demonstrates  the effectiveness of IRD regularization term and the memory replay 
module in maintaining the representation distribution, and without them the representations drift significantly, suffering from the CF problem.

\end{document}